# Bi-stable thin soft robot for in-plane locomotion in narrow space

Xi Wang, Jung-che Chang, Feiran Wang, Dragos Axinte, Xin Dong[1]

*Abstract*—Dielectric elastomer actuators (DEAs), also recognized as artificial muscle, have been widely developed for the soft locomotion robot. With the complaint skeleton and miniaturized dimension, they are well suited for the narrow space inspection. In this work, we propose a novel low profile (1.1mm) and lightweight (1.8g) bi-stable in-plane DEA (Bi-DEA) constructed by supporting a dielectric elastomer onto a flat bi-stable mechanism. It has an amplified displacement and output force compared with the in-plane DEA (I-DEA) without the bi-stable mechanism. Then, the Bi-DEA is applied to a thin soft robot, using three electrostatic adhesive pads (EA-Pads) as anchoring elements. This robot is capable of crawling and climbing to access millimetre-scale narrow gaps. A theoretical model of the bi-stable mechanism and the DEA are presented. The enhanced performance of the Bi-DEA induced by the mechanism is experimentally validated. EA-Pad provides the adhesion between the actuator and the locomotion substrate, allowing crawling and climbing on various surfaces, i.e., paper and acrylic. The thin soft robot has been demonstrated to be capable of crawling through a 4mm narrow gap with a speed up to 3.3mm/s (0.07 body length per second and 2.78 body thickness per second).

*Index Terms*—Thin soft robot, Dielectric elastomer actuators, Bi-stable, Narrow space locomotion, Low profile

## I. INTRODUCTION

Narrow space inspection is commonly seen in human life and industry, especially in scenarios humans cannot access. Robots play a significant role in those fields [1], i.e., continuous robots with small diameters have been developed to access the gas turbine and conduct repair work [2, 3]; A hexapod robot with an embedded machine tool can get into a hazardous environment for maintenance tasks [4]. However, these robots usually have a large volume, complex actuation system, and rigid profile, which will be a limitation in fragile and high-precision scenarios.

Soft robots can overcome those drawbacks with the inherent complaint body, compact profile, lightweight, simple control strategy and reliable human-machine interaction [5]. Many actuation methods have been investigated and applied in locomotion soft robot development. A quadruple pneumatic robot crawled through a 2cm narrow gap using a soft body with a 9mm thickness [6]. The pneumatic chamber with reinforced fibre was capable of controllable deformation [7]. Therefore, an inchworm-inspired soft robot can perform multimodal locomotion of crawling, climbing and transition [8]. Another benefit of the pneumatic actuator is the electronic-free achieved by a buckling-sheet ring oscillator; it also has multimodal movement in various terrains [9]. Most attempts to use the pneumatic robots in narrow space access rely on their complaint body. But, reducing the dimension from centimetre to millimetre scale is a big challenge for pneumatic actuation. Piezoelectric actuated robots have fast responses and more compact volumes. It has been designed into sub-gram with a low profile and can fast-moving 20 body lengths per second [10]. However, it only worked on horizontal locomotion. Alternatively, a microrobot could walk on the vertical and inverted substrate by adding four electrostatic adhesive pads onto a quadrupedal piezoelectric actuated frame [11]. But the dimension still limits its access to millimetre-scale narrow space due to the robot design.

Another actuation option for the soft locomotion robot is the dielectric elastomer actuator (DEA) which is outstanding for its fast response, high energy density and lightweight [12]. Two typical linear DEAs can be applied to soft crawling robots, i. e., cylindrical (spring-roll) linear DEA [13] and saddle-shaped DEA [14, 15]. A pipeline inspection robot consists of a 6mm diameter middle cylindrical DEA as a linear actuator and two shorter DEAs as anchoring feet that can crawl into a pipe [13]. The saddle-shaped DEA worked as a bending actuator and usually used electro-adhesive pads [14] or unidirectional frictions unit [16] to achieve locomotion. A soft wall climbing robot has been demonstrated crawling through a 10mm height narrow gap [14]. With the development of advanced manufacturing technology, multi-layered dielectric elastomers have become a popular topic among researchers. The integrated multi-layers feature can enhance the energy density of the actuator by increasing the output force and reducing the actuation voltage. Based on this improvement, a unimorph multi-layered bending DEA has been designed into a crawling robot [17]. Taking advantage of the amplified vibration of a system in its natural frequency, some researchers proposed soft robotic insects have fast responses and speed actuated at the system's natural frequency [18, 19]. Although the advances in investigating the small dielectric elastomer actuated soft robot, the minimum dimension is still a centimetre-scale, especially regarding the low profile.

An approach to construct millimetre-scale DEA-based soft robots is to utilize the in-plane deformation of DEAs, reducing the space required during actuator morphing. An all-soft, skin-like millimetre-scale structure was proposed with

[1]The research leading to these results has received funding from China Scholarship Council, the Industrial Strategy Challenge Fund delivered by UK Research and managed by EPSRC under Grant Agreement No. EP/R026084/1 and No. EP/P031684/1.

Corresponding Author: Xin Dong (email: xin.dong@nottingham.ac.uk).

Xi Wang, Jung-che Chang, Feiran Wang, Dragos Axinte and Xin Dong are with Department of Engineering, University of Nottingham, NG8 1BB, UK (email: xi.wang1@nottingham.ac.uk; jung-che.chang@nottingham.ac.uk; feiran.wang@nottingham.ac.uk; dragos.axinte@nottingham.ac.uk).

in-plane deformation, capable of locomotion and transportation [20]. However, one limitation was its low velocity of less than 0.12 mm/s (0.001 body lengths per second). In our prior work, we developed a thin DEA-based soft robot utilizing a re-entrant structure that enables in-plane linear deformation, achieving a speed of 2.3 mm/s (0.035 body lengths per second) [21]. One factor limiting the speed was the voltage-induced displacement of the actuator. The negative stiffness bi-stable mechanism is a promising method to enhance the displacement output of the DEA and has been widely studied in cone-shaped DEAs [22, 23], which utilize out-of-plane deformation. Our previous research has demonstrated the feasibility of developing negative stiffness DEAs with in-plane deformation [24], and the thickness can potentially be reduced to millimetre-scale. These advancements pave the way for the development of millimetre-scale thin soft robots with improved speed performance.

This paper presents the design of a novel Bi-stable thin soft robot that can move and access millimetre-scale narrow gaps on horizontal and vertical substrates. The electrostatic adhesive pads (EA-Pads) are used as the anchoring element [14, 25]. Inspired by the out-of-plane cone-shaped DEA tensioned by a biasing mechanism [22, 23, 26], the displacement and the output force of the DEA can be improved by properly balancing the negative stiffness property of the mechanism with the dielectric elastomer. A bi-stable in-plane dielectric elastomer actuator (Bi-DEA) has been presented. It has a thin feature (1.1mm) and amplified displacement and output force compared with the in-plane dielectric elastomer actuator (I-DEA) without the bi-stable mechanism. The rest of the paper is organized as follows. Section II shows the structure of the robot and the design principle of the Bi-DEA. Section III illustrates the fabrication and characterization of the Bi-DEA and the EA-Pad. The experimental results of the robot crawling in narrow spaces on both horizontal and vertical surfaces (paper and acrylic material) have been presented in section IV.

## II. BI-STABLE THIN SOFT ROBOT

### A. Robot Design

As illustrated in Fig. 1a, the bi-stable thin soft robot consists of a Bi-DEA (Fig. 1b) for linear motion and three EA-Pads (Fig. 1c) for anchoring. The front and rear EA-Pads are fixed at the edge of the rectangular frame, while the middle EA-Pad is connected with the shuttle of the bi-stable mechanism. 1-DOF (one-degree-of-freedom) crawling locomotion is realized by alternatively charging Sectors I and II of the DEA (Fig. 1d). The in-plane motion of the Bi-DEA achieves a thickness of less than 1.2mm for the robot. Therefore, it can crawl and access narrow spaces.

### B. Bi-stable in-plane DEA

As illustrated in Fig. 1b, the novel thin Bi-DEA has a dielectric elastomer sandwiched by electrodes on both sides and further bonded onto a bi-stable mechanism. The bi-stable mechanism consists of a central shuttle connected to a rectangular frame through four complaint links. The shuttle and the complaint links divide the electrode into two sections, i.e., Sectors I and II (Fig. 1d). Therefore, the central shuttle can switch between two stable states by applying voltage for Sectors I or II. This bi-stable mechanism with in-plane linear motion can guarantee the thin thickness and improve the voltage-induced displacement and force of the actuator.

An in-plane DEA (I-DEA), as shown in Fig. 2a, and a bi-stable mechanism (Fig. 2b) are proposed to analyze the Bi-DEA displacement and force amplification principle (Fig. 2c). The I-DEA and Bi-DEA have the same configuration, except that the I-DEA does not have the bi-stable mechanism (four complaint links). $F_D$ is the force of the I-DEA to the shuttle when the voltage is applied to Sector I. It has seen linear decreases to zero from the point $a_1$ to $a_3$ where achieves the maximum displacement ($a_3 - a_1$) of the I-DEA. $F_M$ is the force that moving the shuttle from one stable state $a_1$ to the another $a_5$ [27]. For the Bi-DEA in which a bi-stable mechanism is adopted in I-DEA, the output force (shuttle) can be expressed by $F_D - F_M$. The maximum displacement is $a_4 - a_1$ at which point the force goes down to zero ($F_D - F_M = 0$). Therefore, the displacement and the output force of Bi-DEA can be larger than the I-DEA if the force of the dielectric elastomer and the mechanism matches properly, i.e., $a_4 - a_1 > a_3 - a_1$ and $F_D - F_M > F_D$.

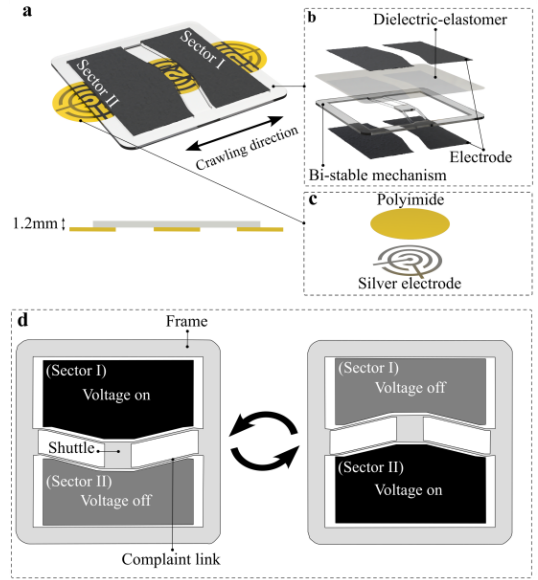

Fig. 1. Concept of Bi-stable thin soft robot. (**a**) 3D model of Bi-stable thin soft robot; (**b**) Explode view of Bi-DEA; (**c**) Explode view of EA-Pad; (**d**) Actuation principle of Bi-DEA.

For the I-DEA with the actuated Sector II, as shown in Fig. 2a, the spring-dashpot system illustrated in Fig. 2d is used to model the response in the Y direction. Based on the non-equilibrium thermodynamic theory [28] and using the Bergstrom-Boyce model [29] to characterize the dielectric elastomer, the voltage-induced stress can be expressed as

$$\sigma_y = \mu_{ye} \frac{\lambda_y^2 \xi_{ye}^{-2} - \lambda_y^{-2} \lambda_x^{-2} \xi_{ye}^2 \xi_{xe}^2}{1 - \frac{\lambda_y^2 \xi_{ye}^{-2} + \lambda_x^2 \xi_{xe}^{-2} + \lambda_y^{-2} \lambda_x^{-2} \xi_{ye}^2 \xi_{xe}^2 - 3}{J_{ye}}} + \mu_y \frac{\lambda_y^2 - \lambda_y^{-2} \lambda_x^{-2}}{1 - \frac{\lambda_y^2 + \lambda_x^2 + \lambda_y^{-2} \lambda_x^{-2} - 3}{J_y}} - \varepsilon E_v^2 \quad (1)$$

where $\lambda_x$ and $\lambda_y$ are the strain of the elastomer in X and Y directions, they are calculated through $\lambda_x = \frac{l_x}{L_x}$ and $\lambda_y = \frac{l_y}{L_y}$, ($L_x$ and $L_y$ are initial length of elastomer in X and Y direction while $l_x$ and $l_y$ represent corresponding length after pre-stretch and actuation). Further, $\lambda_y = \lambda_{ye}\xi_{ye}$; $\mu_y$, $\mu_{ye}$, $J_y$ and $J_{ye}$ are shear modulus and stretch limit of elastomer, as shown in the Fig. 2d; $\varepsilon$ is the permittivity of the dielectric elastomer, $E_v = \frac{U}{l_z}$ ($U$ is the input voltage and $l_z$ is the thickness of the elastomer). Therefore, for a cross-section of $l_x \times l_z$ of Sector II, the output force is

$$F_D = l_x l_z \sigma_y \quad (2)$$

When the shuttle of the bi-stable mechanism moved from one stable state $a_1$ to the another $a_5$, the $F_M$ is comes from the deformation of four complaint links. Each complaint link can be simplified as a fixed-guided beam modelled (Fig. 2e) by the elliptic integral solution [30, 31]. The fixed end is connected with the square frame, and the guided end is fixed with the shuttle. The following equations can describe the relation between reaction force and displacement of the guided end [32]:

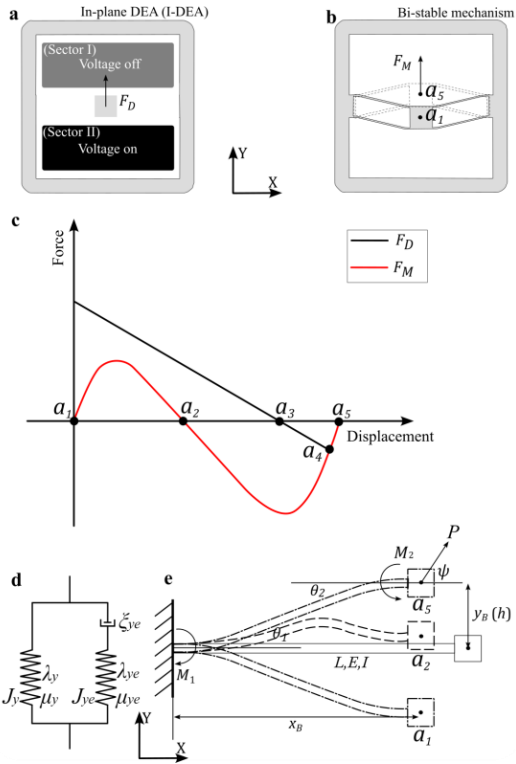

Fig. 2. Principle and analysis of Bi-DEA. (**a**) I-DEA; (**b**) Bi-stable mechanism; (**c**) Force-displacement diagram of DEA and bi-stable mechanism; (**d**) Sketch of the elastomer deformation represented by a spring-dashpot system (Bergstrom-Boyce model [24, 29]); (**e**) Sketch of a fixed-guided beam.

$$\sqrt{\alpha} = F(k,\phi_2) - F(k,\phi_1) \quad (3)$$

$$\frac{y_B}{L} = -\frac{1}{\sqrt{\alpha}}\{2k\cos\psi\,(\cos\phi_1 - \cos\phi_2) + \sin\psi\,[2E(k,\phi_2) - 2E(k,\phi_1) - F(k,\phi_2) + F(k,\phi_1)]\} \quad (4)$$

$$\frac{x_B}{L} = -\frac{1}{\sqrt{\alpha}}\{2k\sin\psi\,(\cos\phi_2 - \cos\phi_1) + \cos\psi\,[2E(k,\phi_2) - 2E(k,\phi_1) - F(k,\phi_2) + F(k,\phi_1)]\} \quad (5)$$

where $\alpha = \frac{PL^2}{EI}$ is the non-dimensional force, $\frac{x_B}{L}$ and $\frac{y_B}{L}$ are the non-dimensional lengths of the guided end of the beam in X and Y direction; $L$, $E$ and $I$ are the length, young's modulus and moment of inertia of the beam; $P$ and $\psi$ are the reaction force and the angle with respect to the X axis; $F(k,\phi)$ and $E(k,\phi)$ are incomplete elliptic integrals of the first and second kind; $k$ is the elliptic modulus, $\phi$ is the amplitude of the elliptic integral and it changes continuously from $\phi_1$ at the fixed end ($\theta_1$) to $\emptyset_2$ at the guided end ($\theta_2$). The $\phi$ also can be calculated by [32]:

$$k\sin\phi = \cos\frac{\psi-\theta}{2} \quad (6)$$

$$M_{1,2} = 2k\sqrt{EIP}\cos\phi_{1,2} \quad (7)$$

where $M_1$ and $M_2$ are the moment of two ends of the beam, $\theta$ is the deformation angle.

As shown in Fig. 2e, the boundary condition is $\theta_1 = \theta_2 = 0$ in the process of the guided end moves from $a_1$ to $a_5$. Therefore, the $\phi$ can be expressed as follow:

$$\sin\phi_{1,2} = \frac{1}{k}\cos\frac{\psi}{2} \quad (8)$$

For the bi-stable mechanism in this paper, the complaint link has gone through two order modes, therefore, $\phi_1$ is the principal solution of Eq (8). For the first order mode, $\phi_2 = \pi - \phi_1$; For the second order mode, $\phi_2 = \phi_1 + 2\pi$.

When the coordinates of the beam's guided end ($x_B, y_B$) are known, the force and moment at the beam's end ($P, M$) can be numerically determined. Starting with initial values for $\psi$ and $k$, a numerical iteration process based on Eqs. (3-5) can be applied to refine the values of $\psi$ and $k$. Then, the load at the guided end can be computed. The reaction force of the shuttle in the Y direction is

$$F_M = 4P\sin\psi \quad (9)$$

### III. FABRICATION AND CHARACTERIZATION

#### A. Fabrication

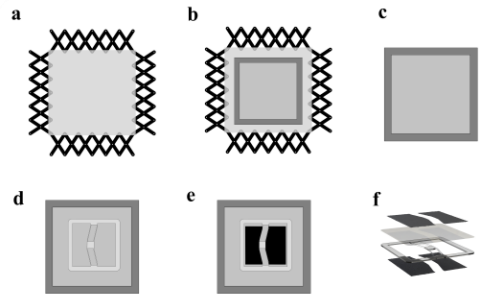

Fig. 3. Fabrication of the Bi-DEA. (**a**) Pre-stretch the elastomer; (**b**) Adhere an acrylic frame onto the pre-stretched elastomer; (**c**) Transfer the pre-stretched elastomer to the acrylic frame; (**d**) Fix the bi-stable mechanism onto the pre-stretched elastomer; (**e**) Paint electrode; (**f**) Explode view of the Bi-DEA.

Fig. 3 shows the fabrication process of the Bi-DEA. A 1mm thick dielectric elastomer (3M VHB 4910) was pre-stretched by 4.5*4.5 times [21] using the fixture in Fig. 3a. In Fig. 3b-c, the pre-stretched elastomer was transferred to a rigid acrylic frame. Then, a bi-stable mechanism was cut from a 1mm thick PETG sheet using the laser cutter (Jindiao Technology Co., Ltd, JD3050). Its shuttle was then manually switched to second stable state ($a_5$, as illustrated in Fig. 2b) and fixed to the elastomer (Fig. 3d). We used multi-walled carbon nanotube as the electrode paints on both sides of the elastomer, as shown in Fig. 3e. Finally, the Bi-DEA was finished by cutting off the acrylic frame (Fig. 3f). Based on this fabrication procedure, the actuator has a 1.1mm thickness and 1.8g weight.

The EA-Pad was fabricated by inkjet printing the electrode onto a polyimide film (diameter: 25mm; thickness: 0.025mm) to the structure, as illustrated in Fig. 1c. The silver electrodes were printed using a Fujifilm Dimatix Materials Printer DMP2850 Series and a 2.4 pL Samba cartridge loaded with the XTPL IJ-36 silver nanoparticle ink. The droplet diameter is 40 µm, and a drop spacing of 20 µm was used to ensure the deposition of continuous layers and pattern uniformity. Six layers were printed on a polyimide substrate, and the substrate was heated to 80 °C by an in-situ resistance heater to pin the droplets. The printed samples are then sintered at 150 °C for an hour to enhance electrical conductivity [33]. This manufacturing method could achieve a thin thickness of 0.028mm and a smooth electrode surface with a sheet resistance of 0.24±0.05 Ω/sq in the pad.

*B. Characterization of Bi-DEA*

Based on the analysis of the working principle of Bi-DEA, the force of the dielectric elastomer and the bi-stable mechanism should be balanced to achieve one stable state in voltage off status. Applying the voltage to one Sector, the voltage-induced force is generated, resulting in the shuttle being moved to the second stable state. As presented in the fabrication of the Bi-DEA, the pre-stretch ratio of VHB was 4.5*4.5, which means the force provided by the elastomer is fixed. Therefore, we need to design the parameters of the bi-stable mechanism to fit $F_M$ with $F_D$, as illustrated in Fig. 2c.

Calculating the bi-stable mechanism based on the mathematical model, Fig. 4a presents the layout and dimension. An 8mm*8mm inner central shuttle is connected to a 50mm*50mm square inner frame through four 0.6mm width flexible links. The central shuttle is biasing from the centre of the frame with a distance of $h$, designed as a variable to explore how the bi-stable property can be optimized. The experimental setup used for this measurement is shown in Fig. 5a. A load cell (Omega LCMFD-10N) was installed on a linear motor (Maxon EPOS 2) and further connected to the shuttle through a lightweight beam. During the measurement, the motor dragged the shuttle at a speed of 0.1mm/s while the load recorded the reaction force. We investigated the force-displacement diagram of the bi-stable mechanism with the $h$ of 4mm and 4.5mm, as shown in Fig. 4b. It was noticed that a larger biasing distance $h$ has a larger displacement and resistant force between two stable states. The error between the simulation and experimental data is attributed to the laser cutting of the compliant link into such a small dimension (0.6 mm), which reduced its stiffness due to the laser cutting process.

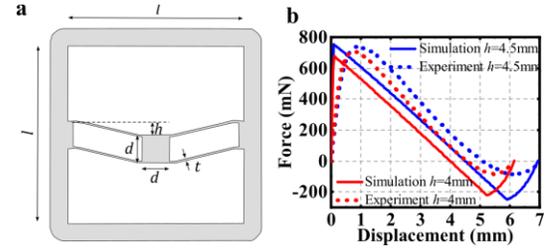

Fig. 4. Characterization of Bi-stable mechanism. (**a**) Layout and dimensions of the Bi-stable mechanism; (**b**) Experimental and simulation diagram of force and displacement of the Bi-stable mechanism.

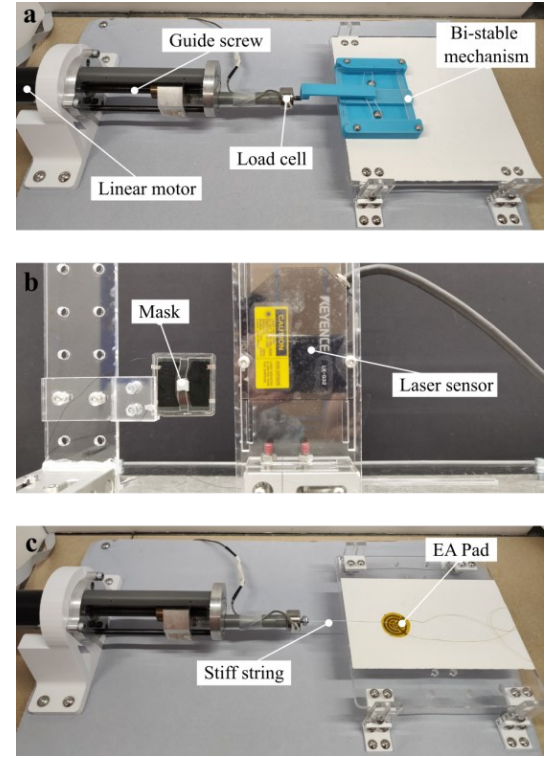

Fig. 5. Experimental setup. (**a**) Bi-stable mechanism and Bi-DEA force measurement; (**b**) Bi-DEA displacement measurement; (**c**) EA pad tangential force measurement.

We characterized the performance of the Bi-DEA and further compared it with the I-DEA to illustrate how the in-plane bi-stable mechanism help to improve the displacement and the output force of the DEA. Fig. 5a-b presents the experimental setup of the characterization. The experimental results of the I-DEA and two types of Bi-DEA ($h$=4mm and $h$=4.5mm) are illustrated in Fig. 6. Applying step signal to Sector II of the DEAs while Sector I remain voltage off, Fig. 6a records the displacement of the central shuttle in different voltages. The I-DEA has a larger displacement than the Bi-DEAs when the voltage is less than 3.5kv. At 4kv, the displacement of Bi-DEA has dramatically increased to 4.35mm ($h$=4mm) and 6.71mm ($h$=4.5mm) from 2.89mm (I-DEA), which is an improvement of 151% and 232%, respectively. This is because the bi-stable mechanism has been switched from one stable state to another, which results in displacement amplification. Fig. 6b is the displacement in different frequencies at 4kv. The square signal was used to actuate Sectors I and II, resulting in the central shuttle switching between two stable states. The displacement amplitude decreased when the frequency increased for all three DEAs. The Bi-DEA ($h$=4mm) has a larger displacement when the frequency rises from 1-2Hz. Furthermore, the Bi-DEA ($h$=4.5mm) has the largest displacement at a lower frequency (0-0.5Hz). Fig. 6c investigates the output force of the DEAs with the displacement increase, in which Sector II was actuated by the step signal (4kv) and no voltage for Sector I. Three DEAs have a similar force value when the displacement is zero due to the same configuration in the measurement (The shuttle was fixed by the load cell. Therefore, the force at zero displacement is determined by the voltage-induced force of Sector II). The I-DEA shows a linear decrease in force when the displacement increases. In contrast, the force of Bi-DEAs has improved when the displacement is larger than 1mm. The maximum force of Bi-DEA ($h$=4mm) is 400mN, while the Bi-DEA ($h$=4.5mm) is 630mN. This comes from the bi-stable mechanism, at which point (~1mm displacement) the central shuttle has been pushed to the second stable state, and the force of the mechanism is reversed.

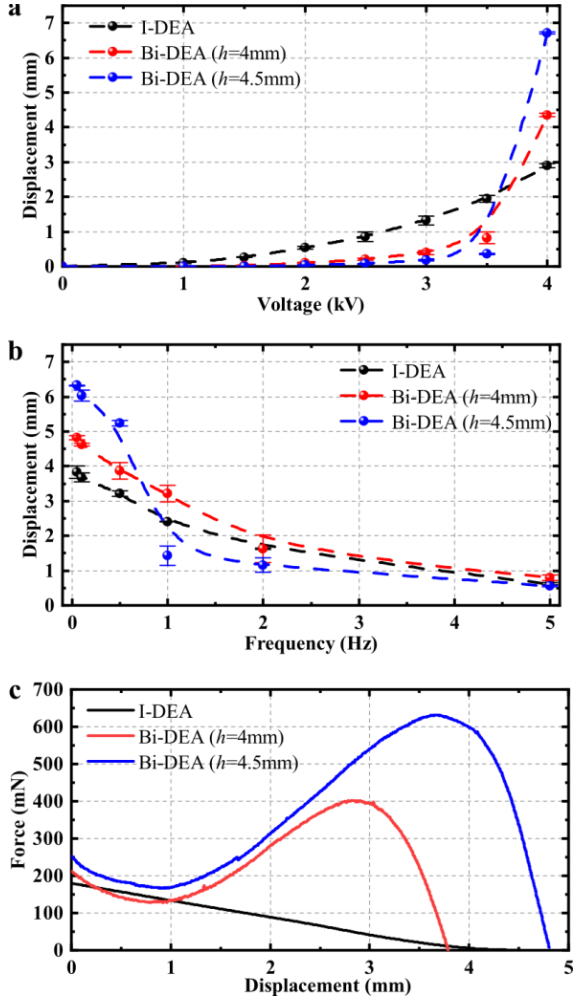

Fig. 6. Characterization of the Bi-DEA. (**a**) Diagram of displacement-voltage (the error bar is the standard deviation of three measurements); (**b**) Diagram of displacement-frequency (the error bar is the standard deviation of three measurements); (**c**) Diagram of force-displacement.

## C. Characterization of EA-Pad

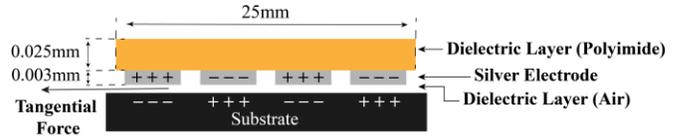

Fig. 7. Working principle of EA-Pad

Fig. 7 illustrates the principle of the EA-Pad with a circular interdigitated geometry. A tangential force is generated between the pad and the substrate with voltage applied. The tangential force of the EA-pad was measured using the equipment shown in Fig. 5c. During the measurement, the motor dragged the pad moving at a constant speed of 0.1mm/s on the substrate. Fig. 8 shows the tangential force of the EA-Pad on paper and acrylic substrate. The value was measured by applying a step signal voltage for the EA-Pad. It can be seen the force increased with the voltage from 0 to 2.5kV. The maximum tangential force is 490mN on the acrylic surface and 549mN on the paper. The differences are caused by the friction coefficient between the pad and acrylic plate is smaller than that of paper.

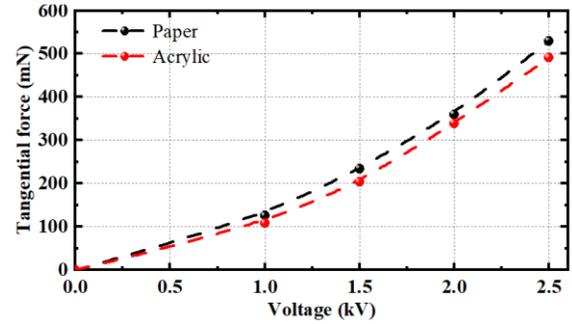

Fig. 8. The tangential force of EA-Pad on paper and acrylic substrate

## IV. EXPERIMENT AND DEMONSTRATION

### A. Gait design

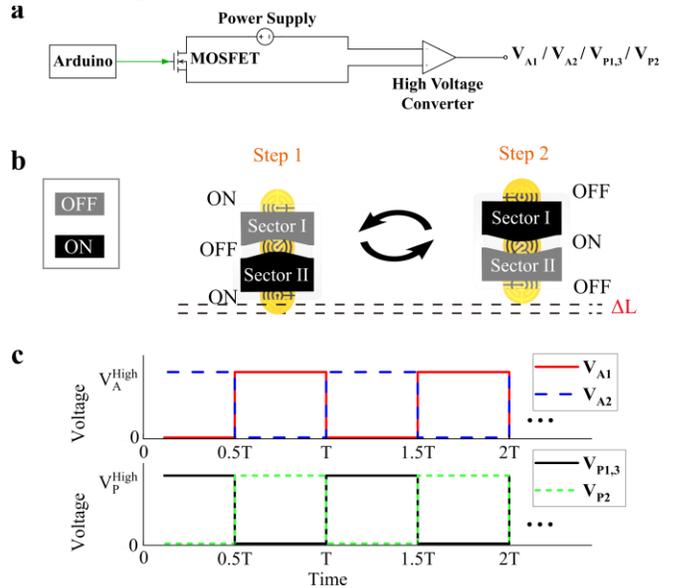

Fig. 9. Control of the bi-stable thin soft robot. (**a**) Schematic of the control system; (**b**) Locomotion gait; (**c**) Sequences of controlled voltages.

Fig. 9a presents the control system for the Bi-stable thin soft robot, the MOSFETs (IRF540N) connected with a power supply and programmed by the microcontroller (Arduino Mega 2560). It can generate a square signal sequence, further amplified by the high voltage converter (XP-Power Q101-5) to actuate the Bi-DEA and EA-Pads. To achieve stable crawling, the motion of the Bi-DEA (controlled by $V_{A1}$ and $V_{A2}$) should be synchronized with the adhesion of the EA-pads (controlled by $V_{P1,3}$ and $V_{P2}$). The crawling of the robot is divided into repeated cycles, T, and each cycle consists of two steps, which is demonstrated in Fig. 9b and Fig. 9c. In the first step (from 0 to 0.5T), the voltage of the front and rear EA-pad, $V_{P1,3}$, increased from 0V to $V_P^{High}$, which caused these pads to adhere to the substrate. At the same time, the applied voltage for the Sector II of DEA, $V_{A2}$, rose from 0V to $V_A^{High}$ and that for Sector I, $V_{A1}$, was at 0V. In this case, the shuttle and the middle EA-Pad is pushed upward by Sector II. In the second step (from 0.5T to T), the voltage applied to the middle pad increased from 0V to $V_P^{High}$ while the $V_{P1,3}$ drop to 0V. This causes the middle pad to adhere to the substrate but the front and rear pads to release from the substrate. With the $V_{A1}$ and $V_{A2}$ changing to $V_A^{High}$ and 0V, respectively, the shuttle moved downward and pushed the robot forward at a distance $\Delta L$. To achieve a higher crawling speed, the Bi-DEA ($h$=4mm) has a higher $\Delta L \times (1/T)$ and will be adopted in Bi-stable thin soft robot.

*B. Crawling*

We demonstrated the robot's ability to crawl on various substrates, using paper and acrylic for the experiments. Fig. 10a shows still images of the robot crawling horizontally on a paper substrate over 30 seconds (see Supplemental Video S1). In this scenario, the robot achieved a maximum speed of 3.33 mm/s (0.07 body length per second and 2.78 body thickness per second) at a frequency of 1 Hz. As illustrated in Fig. 11, the crawling speed increased with frequency from 0.1 Hz to 1 Hz and slightly decreased at 2 Hz. This trend is consistent with the performance of the Bi-DEA shown in Fig. 6b. Fig. 10b demonstrates the robot successfully navigating a 4 mm narrow gap on the acrylic substrate (see Supplemental Video S2).

*C. Vertical crawling*

We further evaluated the robot's vertical climbing performance (see Supplemental Video S3). As shown in Fig. 10c, the robot achieved a maximum vertical climbing speed of 2.38 mm/s on an acrylic substrate (Fig. 11). The vertical speeds were generally lower than the horizontal speeds, particularly at higher frequencies. This is due to the DEA having to overcome its own weight during vertical climbing, which reduces the displacement in each cycle. Additionally, we investigated the robot's potential for confined space inspection. The Bi-DEA generates an output force of up to 400mN, while the EA-Pad can provide an adhesion force of up to 549mN. As demonstrated in Fig. 10d, the robot's payload capacity was tested by successfully climbing vertically on the acrylic substrate while carrying a constant weight of 10 g, which is five times the robot's own weight.

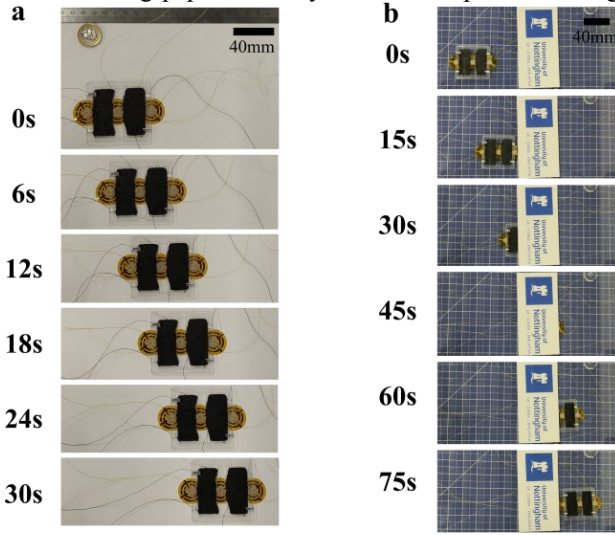
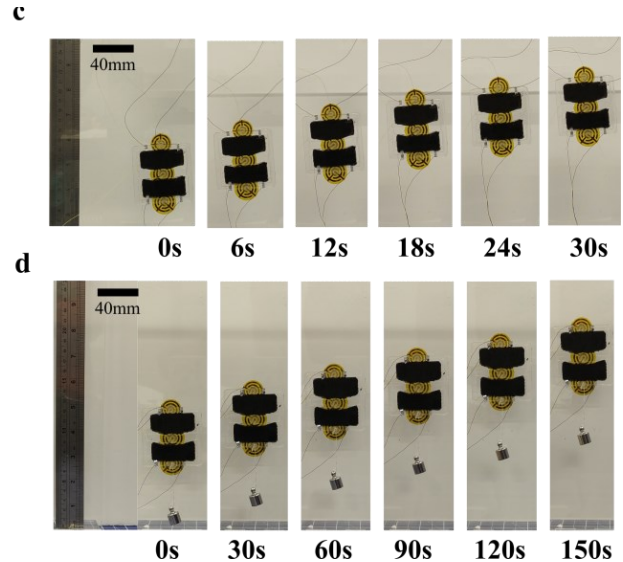

Fig. 10. Locomotion results of the thin soft robot. (**a**) Horizontal crawling on paper; (**b**) Horizontal crawling and accessing a narrow gap on acrylic plate; (**c**) Vertical climbing on acrylic plate; (**d**) Vertical climbing with a 10g payload on an acrylic plate.

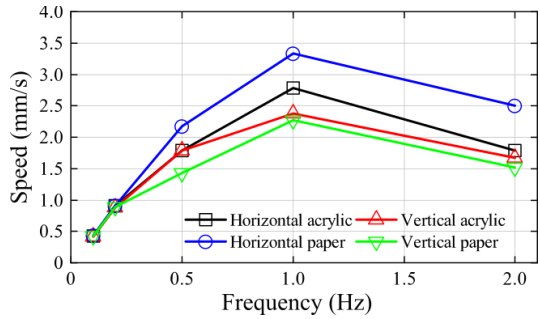

Fig. 11. Locomotion speeds of the bi-stable thin soft robot

V. CONCLUSION

In this paper, we presented a bi-stable thin soft robot with a thickness of 1.2 mm, capable of accessing a 4 mm narrow space both horizontally and vertically on various substrates. The robot consists of two sections: a Bi-DEA and three EA-Pads. The Bi-DEA features amplified displacement and output force by employing an in-plane bi-stable mechanism in the I-DEA. This enhancement increases the displacement to 6.71 mm, representing a 232% improvement over the I-DEA, while the maximum output force rises from 180mN to 630mN. In addition to the enhanced displacement and force properties, the Bi-DEA maintains a low profile (1.1 mm) and lightweight design (1.8 g), making it ideal for miniaturized thin locomotion robots. The three EA-Pads, which serve as adhesive feet, are integrated with the Bi-DEA. The electrodes of the EA-Pads are directly printed onto a 0.025 mm polyimide sheet, resulting in a total thickness of just 0.028 mm. The tangential force between the pad and substrate reaches 549mN on paper and 490mN on acrylic. Thanks to these advantages, we demonstrated the superior performance of the bi-stable thin soft robot compared to the state-of-the-art DEA-based soft robots using EA-Pads as anchoring elements for narrow space access and locomotion (Fig. 12). The robot achieved a maximum speed of 3.33 mm/s, which corresponds to 0.07 body length per second and 2.78 body thickness per second, highlighting its potential for applications in narrow space access and inspection.

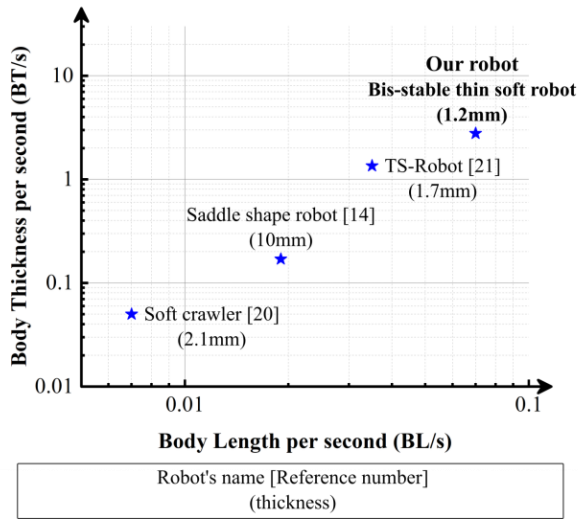

Fig. 12. Performance comparison of our robots with other DEA-based robots with EA-Pad anchoring element. (The speed data for the saddle-shaped robot was calculated based on its crawling demonstration in a confined space with a height of 10mm)

ACKNOWLEDGEMENT

The authors would like to acknowledge Centre for Additive Manufacturing, University of Nottingham for printing of the EA pad that was funded by the Engineering and Physical Sciences Research Council Programme Grant [grant number EP/P031684/1]. The authors also would like to acknowledge the support from Yihan Liu, Junhao Tu and Zonglin Li during the design and fabrication of the Bi-DEA, and Zhichao Wang in preparing the figures.